# Predicting Thermoelectric Power Factor of Bismuth Telluride During Laser Powder Bed Fusion Additive Manufacturing

**Ankita Agarwal*  Tanvi Banerjee*  Joy Gockel†  Saniya LeBlanc‡  Joe Walker§  John Middendorf§**

## Abstract

An additive manufacturing (AM) process, like laser powder bed fusion, allows for the fabrication of objects by spreading and melting powder in layers until a freeform part shape is created. In order to improve the properties of the material involved in the AM process, it is important to predict the material characterization property as a function of the processing conditions. In thermoelectric materials, power factor is a measure of how efficiently the material can convert heat to electricity. While earlier works have predicted the material characterization properties of different thermoelectric materials using various techniques, implementation of machine learning models to predict the power factor of bismuth telluride (Bi2Te3) during the AM process has not been explored. This is important as Bi2Te3 is a standard material for low temperature applications. As a proof of concept, we used data about manufacturing processing parameters involved and in-situ sensor monitoring data collected during AM of Bi2Te3, to train different machine learning models in order to predict its thermoelectric power factor. We implemented supervised machine learning techniques using 80% training and 20% test data and further used the permutation feature importance method to identify important processing parameters and in-situ sensor features which were best at predicting power factor of the material. Ensemble-based methods like random forest, AdaBoost classifier, and bagging classifier performed the best in predicting power factor with the highest accuracy of 90% achieved by the bagging classifier model. Additionally, we found top 15 processing parameters and in-situ sensor features to characterize material manufacturing property like power factor. These features could further be optimized to maximize power factor of the thermoelectric material and improve the quality of the products built using this material.

*Wright State University (agarwal.15@wright.edu, tanvi.banerjee@wright.edu)
†Colorado School of Mines, joygockel@mines.edu
‡The George Washington University, sleblanc@email.gwu.edu
§Open Additive, LLC (jwalker@openadditive.com, jmiddendorf@openadditive.com)

## 1 Introduction

Additive manufacturing (AM) is a process of creating an object by building one layer at a time. The types of AM processes can be classified into different types, one of these is the laser powder bed fusion type (PBF-LB), which can be used to create multiple device components of desired geometry and complexity in one process with minimal waste [1]. In this process, thermoelectric material powder can be melted and fused to create new objects using a laser beam. The steps involved in the manufacturing of these materials determine the overall quality of the product and the efficiency with which heat is converted to electricity [2]. So, predicting the measure of efficiency, known as the power factor, plays an important role in identifying the important parameters involved in the manufacturing process. These parameters and measures could be collected from processing parameters and images captured by in-situ sensors like tomography and polarimetry during the AM process which can be used for material characterization.

The AM process involves processing parameters like laser power, laser travel speed, hatch spacing, layer thickness, and laser focus which influence the output of AM like density, electrical and thermal properties [3, 4, 5, 6]. Additionally, in-situ sensor features and ex-situ material characterization features can also be collected during the AM process to optimize the process [7]. As the processing parameters are modified, the size of the molten pool of material (melt pool) changes. At any given point, this pool is very small (~100 μm). Many passes of the laser and many layers are required to fabricate even a small coupon. Therefore, collecting in-situ data with sensors monitoring the AM process can give valuable insight to the AM processing as additional inputs to predict ex-situ material characterization features like power factor. This is important so as to achieve the desired material properties. Additionally, this can help understand the influence of laser processing on grain morphology and thermoelectric properties which can reveal how PBF-LB processing affects thermoelectric material parts [6]. Optimizing laser processing parameters can also help guide the manufacturing process to build homogenous materials through PBF-LB.

Power factor is equal to the Seebeck coefficient squared multiplied by the electrical conductivity. High performing thermoelectric materials with high Seebeck coefficient and high electrical conductivity but low thermal conductivity is increasingly desirable. The figure of merit Z, which determines the efficiency of thermocouples and thermoelectric generators is directly proportional to the power factor of a thermoelectric material [8]. It is an important material characterization feature which measures the efficiency with which heat is converted to electricity. So, identifying the processing parameters and in-situ sensor features which are useful in predicting power factor can play a critical role in ensuring that materials are built with the optimal thermal and electrical properties, as well as to prevent defects during the PBF-LB process.

PBF-LB has most often been used with metals, ceramics, and polymers and only recently has research been extended to study PBF-LB of thermoelectric materials [2, 4, 5, 9]. While in-situ sensor data has been collected for traditional AM material (e.g.- titanium alloys and nickel superalloys), these data have never been collected for thermoelectric material like bismuth telluride ($Bi_2Te_3$) which is a useful thermoelectric material for refrigeration or portable power generation [10]. Additionally, predicting material properties like power factor using AM processing parameters and in-situ sensor features for this material using machine learning methods have not been done earlier. This can help in modifying the processing parameters to optimize the thermoelectric properties of $Bi_2Te_3$. In this paper, we addressed the following research questions:

- RQ1: Can we build a machine learning model to predict the power factor of thermoelectric material, $Bi_2Te_3$ using processing parameters and in-situ sensor features collected during the AM process?
- RQ2: Which processing parameters and in-situ sensor features play the key role at predicting the power factor of this material during the AM process?

## 2  Related Work

Scientific experiments and simulation methods have been the traditional methods of AM for process parameter optimization which can be time consuming, erroneous and costly [11]. So, earlier researchers have used machine learning methods to optimize processing parameters. With respect to the powder bed fusion AM process of stainless steel 316 L, [12] used laser power and scan speed to predict the melt pool and to construct the process map through gaussian process-based framework. [13] trained random forest model to link the process parameters to pore formation by using data about part orientation, part position, and fraction of recycled powder during the AM process of Inconel 718. Similarly, spreader translation and rotation spread data collected during the AM process of Ti-6A1-4V were used by [14] to train a multilayer perceptron model to construct a process map in order to optimize surface roughness and spreading efficiency for powder bed. To study the influence of support structure parameters on part quality, [15] built decision trees with presence of core support, support density and angle as inputs collected during the AM process. Additionally, ensemble based multi-gene genetic programming was implemented by [16] to achieve desired open porosity values by regulating the process parameters like layer thickness, laser power, and scan speed.

In addition to the processing parameters, in-process monitoring of the layer deposition of the powder using in-situ sensors play a vital role in the quality of the final product and to identify defects in the manufacturing process like distortion, rough surface, cracks, and lack of porosity. Some of these defects may propagate from one layer to the next. [17] detected anomalies of melt tracks through support vector machines and convolutional neural network models using layer wise images of melt pools, plume and spatter captured through high-speed cameras during the AM process of stainless steel 316 L. Similarly, during the AM process of IN 718, high-speed camera and optical microscope captured the morphologies of the melt pool which were used to detect keyhole porosities and balling instabilities using support vector machines [18]. Studies such as [19] detected and located defects due to overheating using K-means clustering on the intensity profile of melt pools captured through high-speed cameras during the AM process of SS316 L. In a different study, analysis of the layer wise surface images before and after powder coating using random forests and support vector machines revealed elevated regions after laser exposure [20]. Self-organizing maps have also been used in previous studies to detect location of pores using thermal profile of the melt pools captured during the AM of Ti-6A1-4V [21].

Artificial intelligence-based data-driven methods have recently been used to discover high performance thermoelectric materials. [22] used machine learning methods to classify the materials into binary classes based on high or low Seebeck coefficients or electrical conductivity based on the features generated using only their molecular formula. Based on the results of machine learning models, they pointed out that machine learning models may misclassify materials that have a relatively low Seebeck coefficient and low thermal conductivity at the same time as poor thermoelectric materials and so predicting figure of merit (Z) or power factor directly through machine learning models is a better choice. [23] predicted thermoelectric performance for layered IV-V-VI semiconductors using high throughput ab initio calculations and machine learning. They generated a dataset from high-throughput ab initio calculations, and developed two neural network models to predict the maximum Z (Z max) and corresponding doping type. Using machine learning models, they were able to identify n-type Pb2Sb2S5 as a potential thermoelectric material with a decent power factor and ultralow thermal conductivity.

While earlier researchers have studied the importance of implementing machine learning methods using processing parameters and monitoring data collected during the AM process for various tasks on different thermoelectric materials, they have not collected in-situ sensor monitoring layer data for $Bi_2Te_3$. Moreover, machine learning models using processing parameters and monitoring layer data to predict how efficiently this thermoelectric material is able to convert heat to electricity during the PBF-LB process have not been implemented. So, we addressed this gap in the literature by predicting a material property like power factor of $Bi_2Te_3$ using processing parameters and in-situ sensor data from tomography and polarimetry collected during its PBF-LB AM process. We chose $Bi_2Te_3$ for this study as it is the only well-established thermoelectric standard reference material for low temperature (up to ≈150 °C) applications [24]. Additionally, it is also the only raw material which is commercially available.

## 3 Methods

### 3.1 Data collection

#### 3.1.1. Laser Powder Bed Fusion Machine

The AM process is controlled by the processing parameters laser power, laser travel speed, hatch spacing, layer thickness, and laser focus. These parameters can be specified for each coupon separately during the PBF-LB manufacturing process. The PBF-LB process works by first spreading powder from the powder stock using the roller over the powder bed. A counter rotating roller re-coater was necessary to help spread the powder into even layers on the build plate due to the irregularities in the particle shape and size. Then the energy source (laser) melts the material on the powder bed in a specified hatch pattern. Next the build platform lowers, and the process is repeated for a given number of layers until the desired thickness was achieved. For this study, coupons were fabricated using a custom built PBF-LB machine at the OpenAdditive facility [1]. Alongside process monitoring sensors were also installed to capture in-situ sensor data. The overall process of the manufacturing process is shown in Figure 1.

#### 3.1.2 Coupon fabrication

$Bi_2Te_3$ coupons were fabricated using PBF-LB using the processing parameter range chosen based prior experience [4]. A total of 220 coupons were successfully manufactured using PBF-LB. So, each coupon was fabricated layer-by-layer. The processing variables used for PBF-LB were laser power, laser speed, hatch spacing, layer thickness, spot size, and scan strategy. Laser powers used for the coupons included 10 W, 12 W, 13 W, 14 W, 16 W, 18 W, 19 W, 20 W, 22 W, 24 W, 25 W, and 30 W. The laser speeds used were 300 mm/s, 350 mm/s, 400 mm/s, 450 mm/s, 500 mm/s, and 700 mm/s. The hatch spacings used included 0.01 mm, 0.02 mm, 0.025 mm, and 0.0375 mm. The layer thickness for the coupons was either 0.1 mm or 0.15 mm. The laser focus determined the spot size of the laser beam on the surface of the specimen. Two spot sizes that were used were 30 µm (focused on the surface) and 257 µm (defocused).

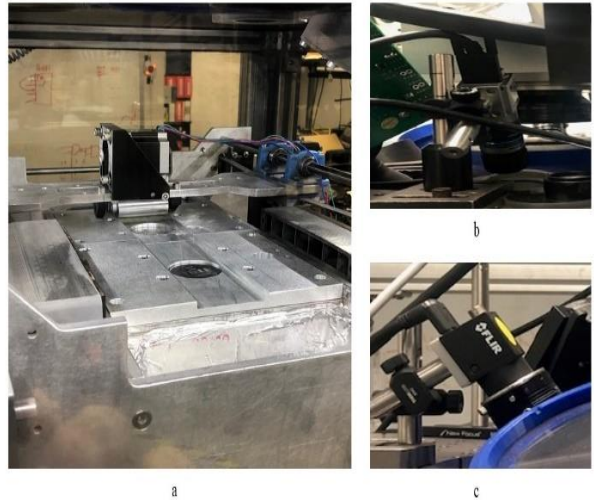

Figure 1: AM process using PBF-LB of $Bi_2Te_3$ showing (a) Laser power bed fusion machine, (b) Tomography sensor, (c) Polarimetry sensor

#### 3.1.3 In-situ sensors

Most approaches to process monitoring in PBF-LB use at least one sensor. We focused on two modalities provided by OpenAdditive. These were polarimetry and thermal tomography. Each modality was captured by an off-axis camera, so we were given a calibration image to compute a homography matrix to calibrate camera perspective of the images into an overhead view. After calibration, each layer image was cropped to the internal area of each coupon. These 15 cropped regions were resampled up or down using Lanczos4 interpolation to 64 x 64 pixels.

#### (a) Thermal Tomography

The thermal tomography was captured by a Basler acA4024-29um camera with a Sony IMX226 CMOS sensor equipped with near-infrared (750-1000 nm) and neutral density filtering lenses. It was configured for long-exposure (250 ms) and low-frame-rate (4 Hz) at a 2000 x 2500 resolution. The sensor had several noisy pixels which required the images to be cleaned using a median window convolution around the hot pixels. Each coupon had a unique processing parameter set that influenced the amount of energy in the welding process. This affected the emission of photons within the camera's operating wavelength spectrum. Sometimes the sensor is not calibrated to handle the full range of emissions which coincidentally washes out texture detail when the sensor response is maxed out by too many emissions. The texture in this sensor modality can be very informative because several phenomena are known to show up in thermal tomography. These include white comets (spatter), dark spots (missing powder), dark lines (recoater streaking or hopping), bright edges (delamination). The thermal tomography data consisted of thermograms from each coupon on each layer.

---
[1] https://openadditive.com/

**(b) Polarimetry**

The polarimetry data was collected by taking two images per layer, one after spreading powder (post spread) and another after melting it (post melt); therefore, no compiling step was necessary. The polarimetry data was captured by a Basler acA4024-29um camera with a Sony IMX226 CMOS sensor that takes 2456 x 2052 resolution monochrome pictures of the build area after spreading a powder layer and after melting it. The post-spread images can reveal the uniformity of the powder spread and whether material from the previous layer is protruding into the next layer. The post-melt images may reveal how the powder spread affect the melted material. They also document each layer such that defects in other process monitoring data may be cross-validated with these data. The polarimetry post-spread data consists of powder spreads from each coupon on each layer while the polarimetry post-melt data consists of melted powder from each coupon on each layer. Both post-spread and post-melt images contained corresponding images for angle of polarization and degree of linear polarization.

**3.2 Image processing and feature engineering**

To implement machine learning models using in-situ sensor data, the raw images collected through these sensors need to be processed and engineered into features before they can be used for building the models. For this purpose, histogram equalization, image filters such as the Weiner filter for deblurring or the median filtering for cleaning were utilized. Then the following features were engineered for each layer of the coupon using Python OpenCV package[2]:

Tomography sensor features: average, median, maximum, minimum, standard deviation of pixel intensity, and surface roughness.

Polarimetry sensor features:
1. Post spread features (features engineered from polarimetry data after the powder layer was spread):
    i. Angle of polarization (AoP): average, median, maximum, minimum, standard deviation of pixel intensity, and surface roughness.
    ii. Degree of linear polarization (DoLP): average, median, maximum, minimum, standard deviation of pixel intensity, and surface roughness.
2. Post melt features (features engineered from polarimetry data after laser scanning was done):
    i. Angle of polarization (AoP): average, median, maximum, minimum, standard deviation of pixel intensity, and surface roughness.
    ii. Degree of linear polarization (DoLP): average, median, maximum, minimum, standard deviation of pixel intensity, and surface roughness.

**3.3 Dataset preparation**

When a coupon was fabricated, in-situ sensor data was collected for each layer of the coupon. The processing parameters remained the same while manufacturing each layer of a particular coupon. Additionally, the ex-situ material characterization output data, power factor at 77°C mW/m K$^2$ was measured for each coupon when it was fabricated. We considered data from 117 coupons for our analysis. We considered each layer of the coupons as a single data point for our analysis. Thus, after performing feature engineering on in-situ sensor data collected for each layer of the coupon, we appended these in-situ sensor features values, corresponding to each layer of the coupon in the dataset with the corresponding value of processing parameters and power factor at 77°C mW/m K$^2$, which were the same for all layers of a particular coupon.

After combining the AM in-situ sensor monitoring data, processing parameters, and power factor to each layer of the coupon, we had a total of 3,157 sample points for our analysis and for building machine learning models. Each layer of a particular coupon had the same value of processing parameters and power factor but different values for in-situ sensor features. The overall framework depicting the various steps involved from data collection, image processing and feature engineering to building predictive machine learning models to predict power factor using processing parameters and in-situ sensor features is shown in Figure 2.

---

[2] https://pypi.org/project/opencv-python/

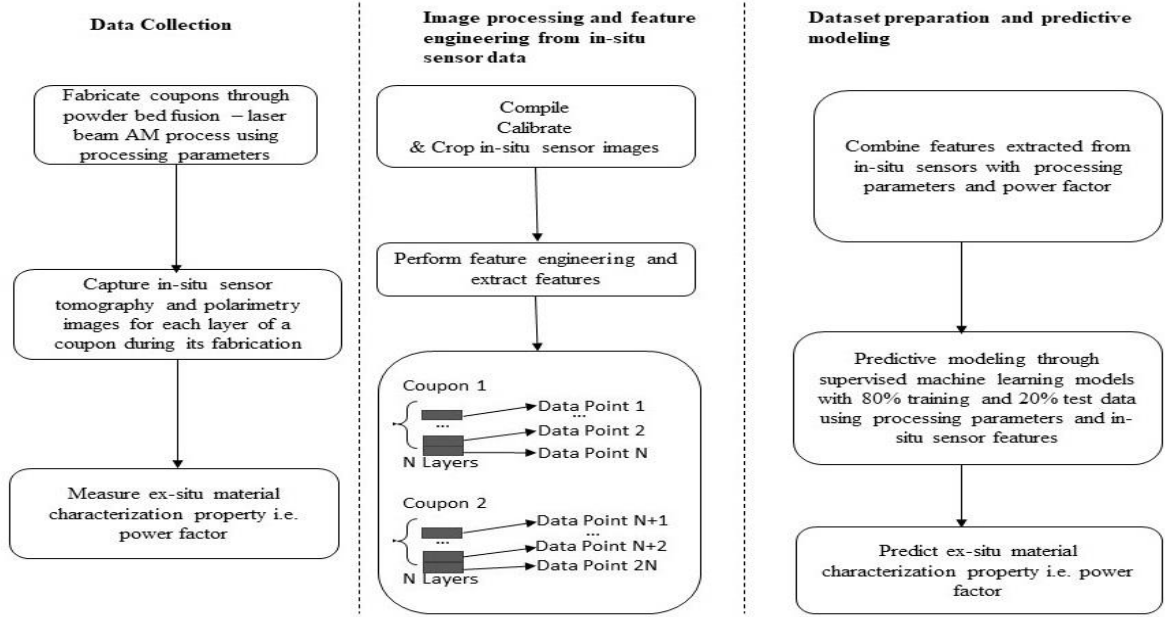

Figure 2: Framework for predicting power factor using processing parameters and in-situ sensor data

**3.4 Machine learning models**

We experimented with four classes of machine learning models in increasing order of complexity: 1) linear models; 2) non-linear kernel-based models; 3) bootstrapping (ensemble) models; and 4) neural networks, to predict a material property known as power factor using processing parameters and in-situ sensor features. As linear models, we implemented naive Bayes, logistic regression, and linear support vector machines (SVM) while polynomial kernel SVM and radial basis function SVM were chosen as non-linear models. We implemented decision tree, random forest, bagging classifier or bootstrap aggregator, and AdaBoost classifier as ensemble models and multilayer perceptron as neural network model. Since the various parameters used as predictors of a target variable come in different ranges, we perform feature scaling (standardization) on processing parameters and in-situ sensor features by normalizing these in the range of 0 and 1. Thereafter, we split our dataset into 80:20 training and test ratio. The random 80% of the dataset was used for training machine learning models while the rest 20% of data was used to test the performance of machine learning models. We implemented these models using the Python scikit-learn library[3].

Naive Bayes classifiers are the probabilistic classifiers based on applying Bayes' theorem with strong 'naive' independence assumptions between the features. Logistic regression estimates the probability of an event occurring based on a linear combination of one or more independent variables. Support vector machines separate the data points on an appropriate hyperplane in an n-dimensional input variable space between the two classes of observations based on the maximal-margin approach. We implemented SVM using three kernel functions. Linear SVM is used for linearly separable data. Polynomial kernel SVM uses a polynomial function to map the data into a higher-dimensional space so that classes can be separated in a hyperplane. Similarly, the radial basis function kernel SVM (RBF kernel SVM) is used when the classes cannot be separated linearly.

Decision trees use a rule-based approach or tree-like model to classify output based on the conditions applied to the independent variables. Random forests are ensemble learning methods for classification that operate by constructing a multitude of decision trees at training time. We set the number of decision trees to be 50 for our model. Similarly, AdaBoost classifiers and bagging classifiers are also ensemble learning methods. In the AdaBoost algorithm, a bunch of models are trained sequentially. We start with the first model which produces some error in the prediction. The next model focuses more on data points that produce errors in the previous model by giving them more weights. This way a series of models are combined for a single prediction. We set the number of estimators to be 50 at which boosting was terminated. Bagging is a bootstrap-aggregated ensemble meta estimator of decision trees where random samples with replacement are generated from the same training set. Then a model is trained on each of these random samples in parallel. Finally, the outcome from all these models are combined. If the target variable is categorical, a majority vote is taken from the results of different models. Otherwise, if the target is numerical, an average of results from all models is calculated. We chose the number of estimators to be 50 to implement bagging. Multilayer perceptron is a type of artificial neural network which consists of three layers of nodes: an input layer, hidden layer and an output layer to predict the output.

---

[3] https://scikit-learn.org/stable/

The power factor values were continuous and so we binned these into two categories high and low based on the median value. Any power factor value less than the median value was categorized as 0 (low power factor) and greater than the median value was categorized as 1 (high power factor). To analyze the performance of a machine learning method, we used accuracy, F1 score, precision, recall, and area under the Receiver Operating Characteristic Curve (ROC AUC) as the performance metrics.

### 3.5 Feature selection

To select the features which were important in predicting the power factor, we used the permutation feature importance technique. A permutation feature importance is defined as the decrease in a model score when a single feature value is randomly shuffled keeping the rest of the features the same. This is done to break the relationship between the feature and the target. By doing so, a drop in the model score indicates how much the model depends on the feature. We implemented the permutation feature importance technique using python scikit-learn library[5].

### 4 Results

The results of the performance of different machine learning methods organized from least to most complex models to predict power factor using processing parameters and in-situ sensor features is shown in Table 1. It can be seen that the ensemble-based methods like random forest, AdaBoost classifier, and bagging performed the best in predicting power factor with the highest accuracy as compared to other machine learning methods.

The top processing parameters and in-situ sensor features selected using the permutation feature importance technique to predict power factor with their

Table 1: Performance metrics for predicting power factor using processing parameters and in-situ features using different machine learning methods organized from least to most complex models. Results of best performing model (Bagging classifier) is highlighted in bold.

| Classification Model | P | R | F1 | AUC | Acc. |
|---|---|---|---|---|---|
| Naive Bayes | 0.72 | 0.72 | 0.72 | 0.77 | 0.72 |
| Logistic Regression | 0.68 | 0.68 | 0.68 | 0.76 | 0.68 |
| Linear SVM | 0.68 | 0.68 | 0.68 | 0.75 | 0.68 |
| Polynomial kernel SVM | 0.68 | 0.67 | 0.67 | 0.75 | 0.67 |
| RBF kernel SVM | 0.68 | 0.68 | 0.68 | 0.75 | 0.68 |
| Decision Tree | 0.88 | 0.88 | 0.88 | 0.97 | 0.88 |
| Random Forest | 0.89 | 0.89 | 0.89 | 0.97 | 0.89 |
| AdaBoost Classifier | 0.88 | 0.88 | 0.88 | 0.88 | 0.88 |
| **Bagging Classifier** | **0.90** | **0.90** | **0.90** | **0.98** | **0.90** |
| Multilayer perceptron | 0.74 | 0.73 | 0.73 | 0.79 | 0.73 |

*Note:* P: Precision, R: Recall, F1: F1 score, AUC: area under the Receiver Operating Characteristic Curve, Acc.: Accuracy

feature importance scores and standard deviation are shown in Table 2. The results of feature selection show that the process parameters were dominant over the in-situ sensor parameters in predicting power factor. Additionally, it was found that the polarimetry sensor features were more important in predicting power factor as compared to the tomography sensor features.

### 5  Discussion

This study played a role in identifying the machine learning models which were best at predicting power factor of thermoelectric material, $Bi_2Te_3$ during the PBF-LB AM of the coupons using this material. Ensemble-based models were found to be the best models to characterize a thermoelectric material property of material like $Bi_2Te_3$. This finding coincides with the earlier studies where the ensemble-based classification scheme was used to detect defects during the AM process using in-situ sensor data in the form of layer wise images [25] and highlights that no individual processing parameter is responsible for the material properties.

Ensemble learning is a suitable way of building machine learning models for sample-based data like the one collected during AM process so that the overall accuracy of the model is less affected by anomalous points that occur when sub-optimal processing parameters are used during the build process. This is important since we do not know apriori what these values are and want to create robust machine learning models that can learn the "weak features" that affect the manufacturing process. Specifically, this occurs when there are more than two or three processing parameters that impact the thermal and electrical properties. The model can be trained either on random samples of data points or the bad data points can

---
[5] https://scikit-learn.org/stable/modules/permutation_importance.html

Table 2: Top 15 important features along with their mean importance score and standard deviation

| S. No. | Feature | Mean score and standard deviation |
|---|---|---|
| 1. | Laser Focus (mm) | 0.142 +/- 0.007 |
| 2. | Power (W) | 0.070 +/- 0.005 |
| 3. | Speed (mm/s) | 0.052 +/- 0.008 |
| 4. | Polarimetry post spread AoP roughness | 0.033 +/- 0.005 |
| 5. | Layer (mm) | 0.030 +/- 0.004 |
| 6. | Polarimetry post melt AoP roughness | 0.029 +/- 0.006 |
| 7. | Hatch (mm) | 0.026 +/- 0.002 |
| 8. | Polarimetry post spread AoP std | 0.017 +/- 0.005 |
| 9. | Tomography roughness | 0.014 +/- 0.003 |
| 10. | Polarimetry post melt DoLP std | 0.013 +/- 0.004 |
| 11. | Polarimetry post melt AoP std | 0.012 +/- 0.005 |
| 12. | Polarimetry post spread DoLP max | 0.012 +/- 0.004 |
| 13. | Tomography median | 0.011 +/- 0.004 |
| 14. | Tomography avg | 0.010 +/- 0.003 |
| 15. | Polarimetry post melt AoP median | 0.010 +/- 0.004 |

*Note:* AoP: Angle of polarization, DoLP: Degree of linear polarization, std: standard deviation, max: maximum, avg: average

be picked over and over again and assigned a higher weight during training. This also reduces the amount of bias and variance of the model [26]. Process parameters play an important role in predicting power factor but as these values remain the same for each layer of the coupon, feature engineering using images collected from in-situ sensors can also reveal the layer-by-layer monitoring data to get insights about the manufacturing process and thermoelectric properties of the material. Roughness is one such feature which is measured using in-situ sensor data. Surface roughness from both tomography and polarimetry data during both spreading and melting process of the powder was identified as an important feature in predicting the power factor of $Bi_2Te_3$. So, if the roughness value exceeds a certain threshold, the manufacturing process may need to be aborted since the properties may be unacceptable and could adversely affect the parts built from that material. The features, tomography median and tomography average identified as important features during the AM process indicated that it is important to reduce the bright or dark spots in the thermal images and finding an optimum laser power to reduce porosity. Finally, the standard deviation of the pixel intensity in the polarimetry data after spreading and melting the powder revealed that the designs or geometry of the parts of the object manufactured using $Bi_2Te_3$ should be modified based on these deviations [27]. These parameters which were best at predicting power factor can further be optimized so as to maximize power factor, enhancing the thermoelectric material's ability to generate electrical power or pump heat.

Designing machine learning models for the AM process has its own limitations and challenges due to a smaller sample size. Techniques like few-shot learning can be implemented [28] in the future to address the lower sample size constraint. Additionally, this study depends on the precision and quality of the images captured by the sensors. Integrating data from multiple sensors can reveal significant patterns in the layer-by-layer manufacturing process.

# 6 Conclusion

To improve the quality of the product manufactured using a thermoelectric material through PBF-LB AM, it is important to be able to predict this material's important characteristic property i.e. power factor. In this study, we implemented machine learning models to predict the power factor of $Bi_2Te_3$ through supervised machine learning techniques using processing parameters involved and in-situ sensor data collected during the manufacturing process. We found that ensemble-based methods, like bagging, performed the best in predicting power factor of $Bi_2Te_3$ with an accuracy of 90%. Additionally, we found the top 15 features from processing parameters and in-situ sensor features to characterize the material manufacturing property like power factor which can further be optimized to maximize the power factor of this thermoelectric material.

**Grant Acknowledgement**

The authors gratefully acknowledge grant support from the U.S. Department of Energy Award Number DE-EE0009097 and National Institutes of Health Award Number R01AT010413.

**Acknowledgement**

The authors acknowledge assistance from Alexander H. Greoger, a MS student and Amanuel Alambo, a PhD student at Wright State University for the initial development of image processing and machine learning algorithms.

## References

[1] K. V. WONG, AND A. HERNANDEZ, *A Review of Additive Manufacturing*, ISRN Mechanical Engineering, (2012), pp. 1–10.

[2] H. ZHANG, D. HOBBIS, G. S. NOLAS, AND S. LEBLANC, *Laser additive manufacturing of powdered bismuth telluride*, Journal of Materials Research, 33 (23) (2018), pp. 4031–4039.

[3] C. OZTAN, B. ŞIŞIK, R. WELCH, AND S. LEBLANC, *Process-microstructure relationship of laser processed thermoelectric material Bi2Te3*, Frontiers in Electronic Materials, 2 (2022).

[4] H. ZHANG, AND S. LEBLANC, *Laser Additive Manufacturing Process Development for Bismuth Telluride Thermoelectric Material*, Journal of Materials Engineering and Performance, 31 (8) (2022), pp. 6196–6204.

[5] C. OZTAN, R. WELCH, AND S. LEBLANC, *Additive Manufacturing of Bulk Thermoelectric Architectures: A Review*, Energies, 15 (9) (2022), 3121.

[6] R. WELCH, D. HOBBIS, A. J. BIRNBAUM, G. NOLAS, AND S. LEBLANC, *Nano- and Micro-Structures Formed during Laser Processing of Selenium Doped Bismuth Telluride*, Advanced Materials Interfaces, 8 (15) (2021), 2100185.

[7] M. GRASSO, A. REMANI, A. DICKINS, B. M. COLOSIMO, AND R. K. LEACH, *In-situ measurement and monitoring methods for metal powder bed fusion: an updated review*, Measurement Science and Technology, 32 (11) (2021), 112001.

[8] H. S. KIM, W. LIU, G. CHEN, C.-W. CHU, AND Z. REN, *Relationship between thermoelectric figure of merit and energy conversion efficiency*, Proceedings of the National Academy of Sciences, 112 (27) (2015), pp. 8205–8210.

[9] A. EL-DESOUKY, M. CARTER, M. A. ANDRE, P. M. BARDET, AND S. LEBLANC, *Rapid processing and assembly of semiconductor thermoelectric materials for energy conversion devices*, Materials Letters, 185 (2016), pp. 598–602.

[10] N. BATISTA, J. CRANDALL, A. EL DESOUKY, S. LEBLANC, S. WANG, AND J. YANG, *Powder Metallurgy Characterization of Thermoelectric Materials for Selective Laser Melting*, in TechConnect Briefs, 4 (2017), pp. 166–169.

[11] C. WANG, X. TAN, E. LIU, AND S. B. TOR, *Process parameter optimization and mechanical properties for additively manufactured stainless steel 316L parts by selective electron beam melting*, Materials & Design, 147 (2018), pp. 157–166.

[12] G. TAPIA, S. KHAIRALLAH, M. MATTHEWS, W. E. KING, AND A. ELWANY, *Gaussian process-based surrogate modeling framework for process planning in laser powder-bed fusion additive manufacturing of 316L stainless steel*, The International Journal of Advanced Manufacturing Technology, 94 (9–12) (2018), pp. 3591–3603.

[13] B. KAPPES, S. MOORTHY, D. DRAKE, H. GEERLINGS, AND A. STEBNER, *Machine Learning to Optimize Additive Manufacturing Parameters for Laser Powder Bed Fusion of Inconel 718*, The Minerals, Metals & Materials Series, (2018), pp. 595–610.

[14] W. ZHANG, A. MEHTA, P. S. DESAI, AND C. F. HIGGS III, *Machine Learning Enabled Powder Spreading Process Map for Metal Additive Manufacturing (AM)*, In International Solid Freeform Fabrication Symposium, Texas, Austin, 2017.

[15] A. DOUARD, C. GRANDVALLET, F. POURROY, AND F. VIGNAT, *An Example of Machine Learning Applied in Additive Manufacturing*, in IEEE International Conference on Industrial Engineering and Engineering Management (IEEM), 2018

[16] A. GARG, J. S. L. LAM, AND M. M. SAVALANI, *A new computational intelligence approach in formulation of functional relationship of open porosity of the additive manufacturing process*, The International Journal of Advanced Manufacturing Technology, 80 (1-4), (2015), pp. 555–565.

[17] Y. ZHANG, G. S. HONG, D. YE, K. ZHU, AND J. Y. H. FUH, *Extraction and evaluation of melt pool, plume and spatter information for powder-bed fusion AM process monitoring*, Materials & Design, 156 (2018), pp. 458–469.

[18] L. SCIME AND J. BEUTH, *Using machine learning to identify in-situ melt pool signatures indicative of flaw formation in a laser powder bed fusion additive manufacturing process*, Additive Manufacturing, 25 (2019), pp. 151–165.

[19] M. GRASSO, V. LAGUZZA, Q. SEMERARO, AND B. M. COLOSIMO, *In-Process Monitoring of Selective Laser Melting: Spatial Detection of Defects Via Image Data Analysis*, Journal of Manufacturing Science and Engineering, 138 (5) (2017).

[20] J. Z. JACOBSMUHLEN, J. ZUR JACOBSMUHLEN, S. KLESZCZYNSKI, G. WITT, AND D. MERHOF, *Detection of elevated regions in surface images from laser beam melting processes*, in IECON 2015 - 41st Annual Conference of the IEEE Industrial Electronics Society, 2015

[21] M. KHANZADEH, S. CHOWDHURY, M. A. TSCHOPP, H. R. DOUDE, M. MARUFUZZAMAN, AND L. BIAN, *In-situ monitoring of melt pool images for porosity prediction in directed energy deposition processes*, IISE Transactions, 51 (5) (2019), pp. 437–455.

[22] D. CHERNYAVSKY, J. VAN DEN BRINK, G. PARK, K. NIELSCH, AND A. THOMAS, *Sustainable Thermoelectric Materials Predicted by Machine Learning*, Advanced Theory and Simulations, 5 (11) (2022), 2200351.

[23] Y. GAN, G. WANG, J. ZHOU, AND Z. SUN, *Prediction of thermoelectric performance for layered IV-V-VI semiconductors by high-throughput ab initio calculations and machine learning*, npj Computational Materials, 7 (1) (2021).

[24] R. R. MONTGOMERY AND J. E. BENKSTEIN, SRM NIST standard reference materials 2018 catalog, 2018

[25] C. GOBERT, E. W. REUTZEL, J. PETRICH, A. R. NASSAR, AND S. PHOHA, *Application of supervised machine learning for defect detection during metallic powder bed fusion additive manufacturing using high resolution imaging*, Addit. Manuf., 21 (2018), pp. 517–528.

[26] X. XIAOYU, J. BENNETT, S. SAHA, Y. LU, J. CAO, W. K. LIU, AND Z. GAN, *Mechanistic data-driven prediction of as-built mechanical properties in metal additive manufacturing*, npj Computational Materials, 7 (1) (2021), pp. 86.

[27] K. ÖZSOY, B. AKSOY, AND O. K. M. SALMAN, *Investigation of the dimensional accuracy using image processing techniques in powder bed fusion*, Proceedings of the Institution of Mechanical Engineers, Part E: Journal of Process Mechanical Engineering, 235 (5) (2021), pp. 1587–1597.

[28] M. R. ZAREI AND M. KOMEILI, *Interpretable Concept-Based Prototypical Networks for Few-Shot Learning*, In 2022 IEEE International Conference on Image Processing (ICIP), Bordeaux, France, 2022.